\documentclass[10pt,twocolumn,letterpaper]{article}

\usepackage{wacv}
\usepackage{graphicx}
\usepackage{amsmath}
\usepackage{amssymb}
\usepackage{booktabs}
\usepackage{multirow}

\usepackage[pagebackref,breaklinks,colorlinks]{hyperref}

\usepackage[capitalize]{cleveref}
\crefname{section}{Sec.}{Secs.}
\Crefname{section}{Section}{Sections}
\Crefname{table}{Table}{Tables}
\crefname{table}{Tab.}{Tabs.}

\begin{document}

\title{
An Improved Method for Personalizing Diffusion Models}


\author{Yan Zeng\textsuperscript{1}
\and
Masanori Suganuma\textsuperscript{1,2}
\and
Takayuki Okatani\textsuperscript{1,2}
\and
{\textsuperscript{1}Graduate School of Information Sciences, Tohoku University
\textsuperscript{2}RIKEN Center for AIP}\\
{\tt\small {yan, suganuma, okatani}@vision.is.tohoku.ac.jp}
}
\maketitle

\begin{abstract}
Diffusion models have demonstrated impressive image generation capabilities. Personalized approaches, such as textual inversion and DreamBooth, enhance model individualization using specific images. These methods enable generating images of specific objects based on diverse textual contexts. Our proposed approach aims to retain the model's original knowledge during new information integration, resulting in superior outcomes while necessitating less training time compared to DreamBooth and textual inversion.
\end{abstract}

\section{Introduction}
\label{sec:intro}

In recent years, with the advent of deep generative models, particularly diffusion models \cite{ho2020ddpm}, it has become possible to generate high-quality images of a wide variety of scenes and objects. Furthermore, text prompts make it possible to control the content of the generated images. In text-to-image synthesis, the model consists of a text encoder, which maps a text prompt into its embedding space, and a diffusion model, which takes the resulting embedding as an additional input to generate an image of the desired scene.

Personalization is a specific application of this technology. This entails the ability to generate images of specific objects (such as a particular bag or one's own pet) based on a few example images of those objects. The challenge lies in generating images of the specific objects rather than those of generic bags or dogs, while matching the examples down to minute details. Similar to regular text-to-image synthesis, personalization often requires prompt-based control over the context of the generated image, such as the background and the pose of the object.

Several approaches have been proposed, all of which presuppose the existence of a pre-trained text-based image generation model. One approach, exemplified by textual inversion \cite{gal2022textual}, involves optimizing the embedding of the prompt text. Generally speaking, expressing the desired specific object in words can be difficult, owing to the great diversity of objects and the relative insufficiency of the expressive power of language, no matter how much rhetoric is employed. Therefore, the idea is to create a new word or phrase to represent the specific object in the text embedding space. More specifically, this approach involves optimizing the text embedding that represents the given specific object from its image examples.

However, it is impossible to create an image that faithfully reproduces the specific object using this method. Due to the relative limitations in the expressive power of the linguistic space, it is likely that there does not exist an embedding in the text embedding space that can represent the specific object itself, including its minute details.

Another method, exemplified by DreamBooth \cite{huggingface2022Dreambooth}, seeks to optimize the image generation model itself. This approach involves specifying an undefined term corresponding to the specific object, and then retraining (fine-tuning) the model so that when that term is input into the generation model, the specific object is generated. For instance, in the case of one's pet dog, the model is trained to generate the given image in response to a prompt such as ``a photo of [xxx] dog.'' However, retraining the model with only a few examples can result in the forgetting of previously learned content. Faithfully generating images of the specific object requires many weight updates, but a larger number of updates increases the risk of forgetting. To address this issue, DreamBooth simultaneously trains the model on a diverse collection of class images. Specifically, in the case of a dog, a diverse collection of dog images is used, and the model is required to generate these images in response to a prompt such as ``a photo of a dog.'' To fulfill this objective, a loss called the prior preservation loss is added to the loss for learning the specific object from its few images, serving as a regularization term. This method enables much more faithful image generation than textual inversion.

However, even with DreamBooth, the fidelity of the generated images is often imperfect. Additionally, the method incurs substantial computational costs. Evaluating the prior preservation loss is computationally expensive, and training the model to generate a single specific object requires a considerable amount of time.

Naturally, a combination of both approaches is conceivable. Imagic \cite{kawar2023imagic} represents a method that integrates both. However, Imagic targets the problem of editing a single image through text rather than personalization. It is not designed to learn the characteristics of a specific object from multiple images.

\section{Related Work}

\subsection{Text-to-Image Synthesis}

With advancements in multi-modal models and large language models, the application of generative models to these frameworks has made text-to-image tasks feasible. In recent years, text-to-image generation models have undergone rapid evolution. There are models based on autoregressive architectures, such as DALL-E\cite{ramesh2021dalle} and Parti\cite{yu2022parti}, as well as those trained using GANs, such as Lafite\cite{zhou2021lafite}. Text-to-image models based on diffusion models have also garnered significant attention. DALL-E2\cite{ramesh2022dalle2} utilizes CLIP\cite{radford2021clip} to transform textual descriptions into image embeddings, which are then decoded to generate images that align with the text. Imagen\cite{saharia2022imagen} introduces a cascaded architecture that initially generates images at a resolution of $64 \times 64$ and subsequently employs a two-stage text-conditioned super-resolution diffusion model to upscale them. Stable Diffusion\cite{Rombach_2022_ldm} is among the earliest open-source text-to-image diffusion models. Unlike other diffusion models, it does not operate directly on pixel images. Instead, it first compresses images into low-dimensional representations and then performs training in the latent space.

While text-to-image models exhibit a high level of semantic consistency between images and text descriptions, they still face limitations due to the inherent constraints of textual descriptions. Particularly, when dealing with objects that possess intricate details, the model may struggle to accurately convey the nuances, resulting in generated images that do not align with expectations.

\subsection{Personalization}

The concept of personalized image generation within the diffusion model framework was first introduced by DreamBooth\cite{ruiz2023dreambooth}. In their paper, the authors refer to this concept as ``subject-driven generation.'' Subsequent papers refer to this task as ``personalization.'' Han et al. \cite{han2023hiper} use highly personalized (HiPer) text embeddings to replace uninformative embeddings and optimize only the HiPer embeddings. During inference, the learned HiPer embeddings replace the final part of the new embeddings, enabling the generation of object images based on different descriptions. Another study employed apprenticeship learning\cite{chen2023suti}. Initially, the diffusion model was fine-tuned to produce expert models. Subsequently, a dataset was curated using imaginary captions proposed by PaLM, along with images generated by the expert model based on these captions. Finally, the apprentice model was trained using this dataset. Kumari et al. \cite{kumari2022customdiffusion} suggested training only the parameters of the cross-attention layers in the U-Net\footnote{The main part of a diffusion model is composed of ResNet blocks, self-attention blocks, and cross-attention blocks in a U-Net architecture.} while freezing the others. They also introduced the idea of training a single model for multiple concepts.

\section{Personalization Approaches}
\label{preliminaries}

The task of personalization is defined as follows: given a collection of images of a specific target object, we aim to produce new images of that object based on input prompts while preserving its detailed features. Subsequently, we review two key studies on the task, highlight their limitations, and introduce our approach.

\subsection{Existing Approaches}

\subsubsection{Textual Inversion}

\begin{figure}[t]
 \centering
  \includegraphics[width=0.9\linewidth]{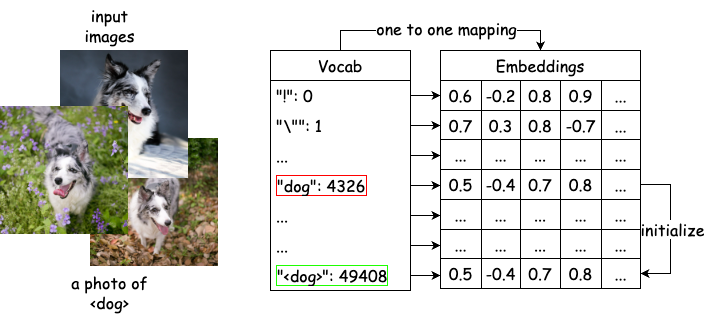}
 \caption{Illustration of textual inversion \cite{gal2022textual}. Given a few sample images of a specific target object, it optimizes the embedding $v_*$ of a newly introduced word/token $S_*$ for the object, enabling the personalization of a pre-trained diffusion model. The embedding $v_*$ is initialized using the embedding of a general class (e.g., ``dog'') of the target object.}
 \label{ti_diagram}
\end{figure}

Textual inversion associates features of a target object with a new word or token by refining its corresponding embedding values, as shown in Figure \ref{ti_diagram}. 

Consider $y$ as an example input text prompt, such as ``a photo of $S_*$'', where $S_*$ represents a novel word designating the target object. This prompt is initially transformed by a tokenizer into a series of word or sub-word indices using a predefined vocabulary that now includes $S_*$ as a fresh entry. Each of these tokens is then linked to an embedding vector via a look-up table, creating a sequence of embedding vectors. The embedding vector associated with the new token for $S_*$ is denoted as $v_*$, which is initialized using the embedding vector of the target object's general class, for instance, `dog'.

The embedding vector $v_*$ is fine-tuned using the same objective function employed in latent diffusion models as stated in \cite{Rombach_2022_ldm}:
\begin{equation}\label{eqn_ldm}
    L_\mathrm{LDM} = \mathbb{E}_{z\sim\mathcal{E}(x), y, {\epsilon}\sim\mathcal{N}(0, I), t} 
    \Big[\|{\epsilon} - {\epsilon}_\theta(z_t, t, \tau_\theta(y))\|_2^2 \Big].
\end{equation}
This loss function measures the L2 distance between the noise ${\epsilon}$, drawn from a standard normal distribution, and the noise ${\epsilon}_\theta$ predicted by the model. Here, $z_t$ denotes the noisy low-dimensional representation of a training image of the target object at timestep $t$. The function $\tau_\theta$ is a pretrained text encoder, whose output $\tau_\theta(y)$ contains the target parameter $v_*$. Meanwhile, the model's parameters and the embeddings of other tokens in the vocabulary are frozen.

One benefit of textual inversion is that the model preserves all prior knowledge because only the embedding of the target object's token undergoes optimization. However, textual inversion often faces challenges in producing correct images of the target object. Capturing the intricate details of the target object using the embedding of a single token is difficult, as the model's embedding space is not designed for this purpose.

\subsubsection{DreamBooth}
\label{3.1.2}

\begin{figure}[t]
 \centering
  \includegraphics[width=0.9\linewidth]{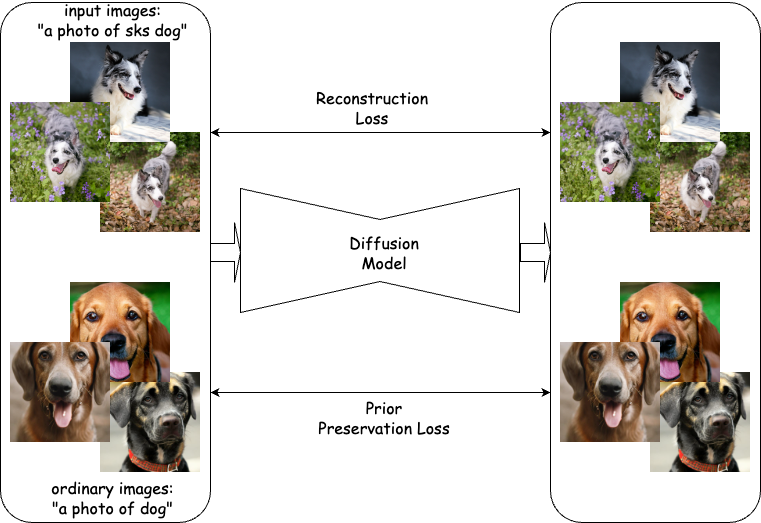}
 \caption{Illustration of DreamBooth \cite{ruiz2023dreambooth}. Given a few sample images of a target object, it fine-tunes a diffusion model by minimizing the sum of two loss functions. The reconstruction loss measures the difference between the generated images and the sample images of the target object. The prior preservation loss measures the reconstruction error for common-class objects, aiming to prevent the model from forgetting how to generate their images.}

 \label{db_diagram}
\end{figure}

DreamBooth adopts an alternative approach. It fine-tunes a pre-trained model using sample images of the target object, as illustrated in Figure \ref{db_diagram}. Contrary to textual inversion, DreamBooth does not add a new token to the vocabulary. The original paper \cite{ruiz2023dreambooth} mentions that infrequently used tokens in the vocabulary are initially identified and then inverted back to the text space; the identifier $S_*$ is finally defined as a character sequence bounded by one to three decoded tokens\footnote{Since DreamBooth did not publicly release its code, we adopt the implementation by Hugging Face and use ``sks'' as an identifier, which is also a rare token.} 

Since DreamBooth has not published its code, it is unclear how the rarity of tokens is defined and how the selected identifier performs. Unlike the more complex setting in DreamBooth, `sks', a rare word in the English dictionary, was selected as the identifier in the earliest implementation\footnote{Implementation of DreamBooth with Stable Diffusion: \url{https://github.com/XavierXiao/Dreambooth-Stable-Diffusion}}. This choice has been maintained in the widely adopted Hugging Face codebase, and our experiments follow the same setup.

Fine-tuning the model using a limited number of images can result in significant overfitting. Additionally, this can cause ``language drift,'' where the model loses its ability to generate images of typical objects. DreamBooth addresses this problem by introducing a `prior preservation loss' as a regularization term. The augmented loss is given as follows:
\begin{multline}
    \mathbb{E}_{x, c, {\epsilon}, {\epsilon}',t} 
    \Big[\|\hat{x}_\theta(\alpha_tx+\sigma_t{\epsilon},c)-x\|_2^2\\+\lambda \|\hat{x}_\theta(\alpha_{t'}x_{pr}+\sigma_{t'}{\epsilon}',c_{pr})-x_{pr}\|_2^2\Big].
    \label{eqn:df}
\end{multline}

The first term corresponds to the original loss function of the diffusion model, which measures the L2 distance between the predicted image\footnote{The backbone of DreamBooth is Imagen, which predicts the original image. Stable Diffusion predicts the noise added to the original image.} $\hat{x}_\theta$ and the original image $x$ that contains the target object. The second term is the prior preservation loss. It measures the L2 distance between the predicted image $\hat{x}_\theta$ and the original image $x_{pr}$ for common-class objects. This necessitates the generation of a batch of images using a pre-trained diffusion model based on prompts that include common-class names.

While DreamBooth works much better than textual inversion and sometimes yields satisfactory results, it has its own limitations. The original paper \cite{ruiz2023dreambooth} mentions several issues, such as incorrect context synthesis (i.e., not being able to accurately generate the prompted context), context-appearance entanglement (i.e., the appearance of the target object changes due to the prompted context), and overfitting (i.e., generating images that are too similar to the input samples). 

In addition to those mentioned in the paper, 
DreamBooth tends to suffer from the following problems.

\smallskip
\noindent
\textbf{Compromised training efficiency} The computation of the prior preservation loss necessitates the prior generation of 1,000 ordinary-class images. This step can take even longer than the training itself, effectively doubling or tripling the training time. This significantly reduces training efficiency.

\begin{figure}[t]
 \centering
  \includegraphics[width=0.9\linewidth]{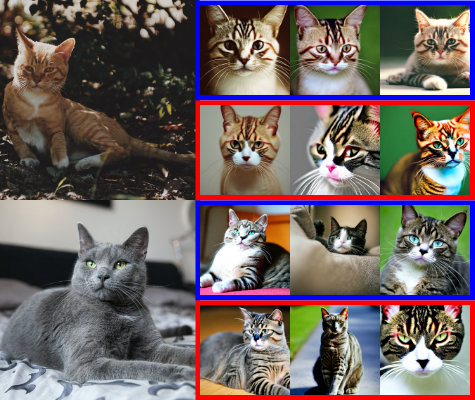}
 \caption{\textbf{Blurriness and diminished realism.} One of the input sample images is shown on the left and the generated results are on the right. The blue boxes indicate the results of our method and the red boxes indicate those of DreamBooth. The prompt is ``a photo of cat.'' All results are taken from the 200-step checkpoints.}
 \label{priorknowledge}
\end{figure}

\smallskip
\noindent
\textbf{Lower quality image generation for common-class objects} 
Across certain datasets, the utilization of the prior preservation loss causes
discernible degradation in the quality of images of ordinary-class objects; see Figure \ref{priorknowledge}.

\smallskip
\noindent
\textbf{Artifacts in generated images} 
It is reported in \cite{huggingface2022Dreambooth} that excessive training in DreamBooth can lead to the emergence of color artifacts in the generated images, which aligns with our findings; see Figure \ref{overtraining}. 

\begin{figure*}[t]
\begin{center}
\includegraphics[width=0.8\linewidth]{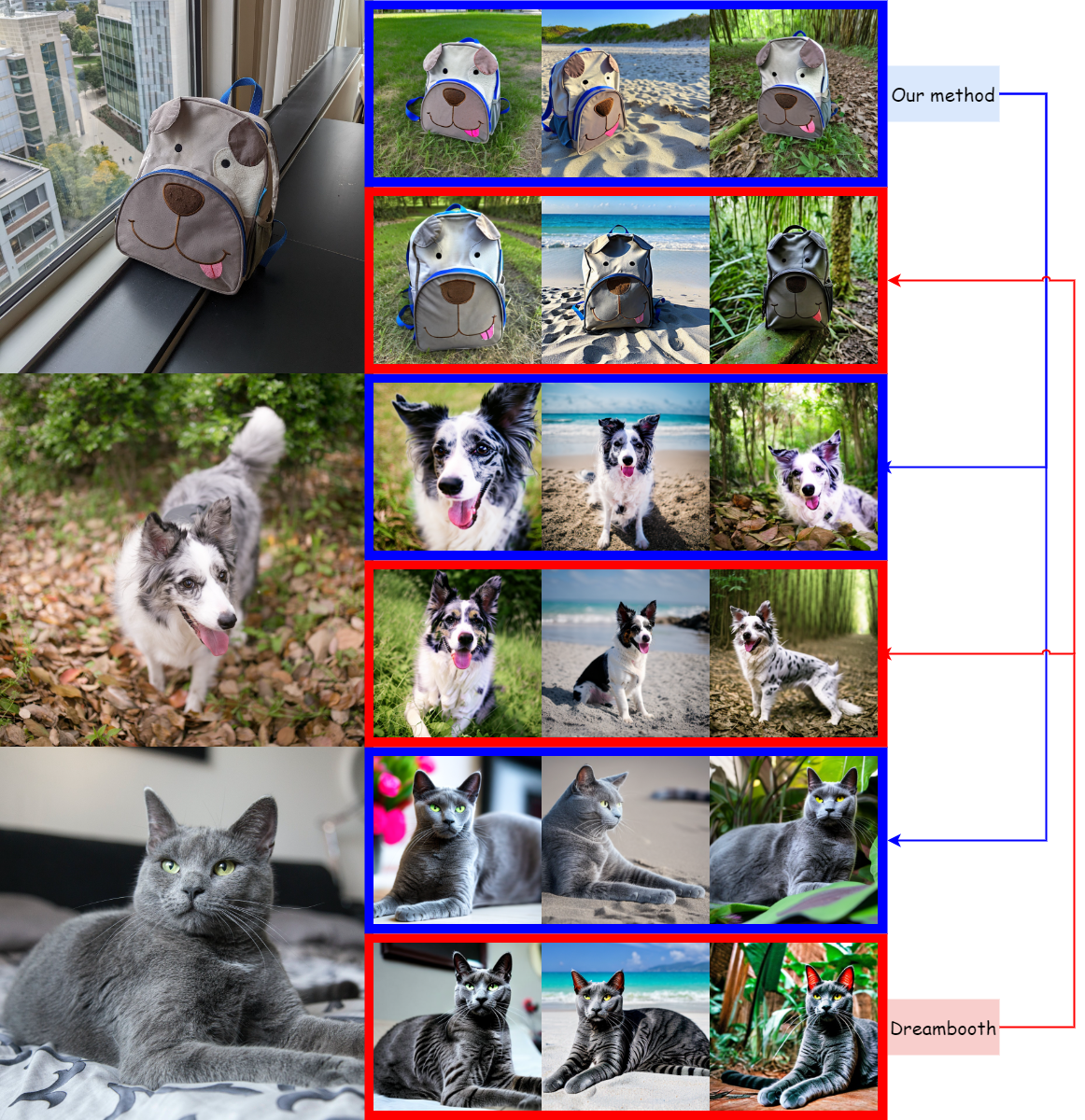}
\end{center}
\caption{
Results of personalization with DreamBooth and our method. 
Reconstruction of `backpack', `dog', and `cat' subject instances from the dataset in \cite{ruiz2023dreambooth}. The results from the best-performing checkpoints are selected and shown here. The images enclosed by blue boxes are the results of our method, with the prompts used from left to right being ``a photo of $\langle rare \rangle$ backpack'', ``a photo of $\langle rare \rangle$ backpack on the beach'', and ``a photo of $\langle rare \rangle$ backpack in the jungle''. The red bounding boxes show DreamBooth's results, with the prompts used from left to right being ``a photo of sks backpack'', ``a photo of sks backpack on the beach'', and ``a photo of sks backpack in the jungle''. As mentioned in \ref{3.1.2}, we adopt the standard Hugging Face setting that uses `sks' as the identifier.}
\label{reconstruction}
\end{figure*}

\medskip
While the prior preservation loss addresses overfitting and language drift, it also introduces new problems. These problems become more pronounced as training time increases. In another paper \cite{zhao2023recipe}, a weight-constrained loss was introduced to slow down the rate of parameter updates. However, these regularization terms are insufficient to address the above issues.

\subsection{Proposed Two-Stage Optimization}

We present an approach that combines textual inversion and DreamBooth. While combining the two is a straightforward idea, their proper integration addresses the issues of each method, as shown by our experimental results. 

Specifically, the proposed approach comprises two stages. In the first stage, a token embedding is optimized similarly to textual inversion. In the second stage, the diffusion model is fine-tuned similarly to DreamBooth, but without the prior preservation loss. The two stages differ from textual inversion and DreamBooth, as explained below. 

In the first stage, we incorporate a new token ``$\langle rare \rangle$,'' which is used as an adjective modifying a noun that represents the target object, e.g., 
\begin{itemize}
\item[]``a photo of $\langle rare \rangle$ backpack.'' 
\end{itemize}
Note that this differs from textual inversion, which utilizes a new word $S_*$ as a {\em noun} representing the target object. We optimize the embedding $v_\mathrm{rare}$ of ``$\langle rare \rangle$'' by
\begin{equation}\label{eqn_stage1}
\min_{v_\mathrm{rare}}\mathbb{E}_{z\sim\mathcal{E}(x), y, {\epsilon}\sim\mathcal{N}(0, I), t} 
    \Big[\|{\epsilon} - {\epsilon}_\Theta(z_t, t, \tau_\theta(y))\|_2^2 \Big],
\end{equation}
with $v_\mathrm{rare}$ initialized with the embedding of the word ``rare''. 

Another difference from textual inversion is that we aim to make ``$\langle rare \rangle$'' represent the appearance of the target object only {\em coarsely}. Thus, we update $v_\mathrm{rare}$ for only about 100 steps. Note that textual inversion aims to make the new token represent the target object as accurately as possible, requiring 3,000--5,000 update steps. 

In the second stage, we fine-tune the parameters $\Theta$ of the diffusion model by
\begin{equation}\label{eqn_stage2}
    \min_{\Theta}\mathbb{E}_{z\sim\mathcal{E}(x), y, {\epsilon}\sim\mathcal{N}(0, I), t} 
    \Big[\|{\epsilon} - {\epsilon}_\Theta(z_t, t, \tau_\theta(y))\|_2^2 \Big].
\end{equation}

This loss uses the input images of the target object and the above prompt containing ``$\langle rare \rangle$.'' Its embedding $v_\mathrm{rare}$, optimized in the first stage, is frozen in this stage; the parameter $\theta$ of the text encoder is also frozen. Note that we do not use the prior preservation loss here. The aim is to train the model conditioned on the token embedding $v_\mathrm{rare}$ learned in the first stage so that it can represent the detailed appearance of the target object. 

There is another difference from DreamBooth: our model typically achieves its best results after only 200--400 fine-tuning steps. In contrast, DreamBooth requires 1,000 parameter-update steps. It also requires the generation of an additional 200 to 1,000 images of common-class objects for the prior preservation loss.

To summarize, our method offers two advantages over existing methods. First, it enhances the quality of generated images, often significantly, as demonstrated in the subsequent section. Second, it considerably reduces training time. Fewer parameter updates also decrease the chances of overfitting and language drift. The removal of the prior preservation loss mitigates potential complications. Together, these changes lead to superior image generation quality.

It should be noted that the approach employed by Imagic \cite{kawar2023imagic} has some similarity to our method; it employs a similar two-stage approach of optimizing the embedding of a prompt followed by optimizing the model. However, Imagic does not explicitly extract the features of the target object; a separate model needs to be trained for each different editing prompt. Our method allows users to freely modify the image based on different text inputs, providing more flexibility in editing.

\section{Experimental Results}
\label{experiment}

\subsection{Experimental Configuration}

We experimentally evaluate the proposed method using the dataset introduced in \cite{ruiz2023dreambooth}. The dataset consists of 30 sets of images, each featuring four to six images containing the same specific object. For the base text-to-image synthesis model, we use the pre-trained latent diffusion model ``stabilityai/stable-diffusion-2-1-base,'' available from Hugging Face: \url{https://huggingface.co/stabilityai/stable-diffusion-2-1-base}.

As explained above, the proposed method consists of two stages. The parameters are updated for 100 iterations with a learning rate of $5 \times 10^{-4}$ in the first stage, and 800 iterations with a learning rate of $5 \times 10^{-6}$ in the second stage. We save a checkpoint every 200 steps in the second stage for evaluation. 

\subsection{Quantitative Evaluation}

To evaluate personalization methods, it is essential to assess how well the model preserves the details of the target object in the generated images. However, designing a metric that aligns with human judgment to measure the similarity of objects in two images is a challenging task.

Since designing such a metric remains an open problem, we follow previous studies \cite{ruiz2023dreambooth} for the comparative evaluation of the proposed and existing methods. 

Specifically, we employ the CLIP score and the DINO score as evaluation metrics; see \cite{dino,radford2021clip} for their definitions. The evaluation process typically unfolds as follows. Initially, we employ the fine-tuned model to produce images of the target object. Subsequently, both the generated and input sample images of the target object are encoded into their respective embeddings, either through the CLIP image encoder or the DINO vision transformer. We then compute the cosine similarity between these embeddings to gauge the resemblance between the generated and sample images. For this evaluation, we utilize two prompting methods: one sourced from DreamBooth \cite{ruiz2023dreambooth} and the other from textual inversion \cite{gal2022textual}.

The first method from DreamBooth uses 25 diverse prompts\footnote{The complete prompts are available here: \url{https://github.com/google/dreambooth/blob/main/dataset/prompts_and_classes.txt}} with additional descriptions. For each prompt, we generate four images, yielding 100 images in total. This aims to test whether the model can produce diverse images. Some prompts might change the object's shape or color, e.g., ``a red $\langle rare \rangle$ dog'' or ``a cube-shaped $\langle rare \rangle$ dog,'' which can lead to a decrease in the CLIP score.

The second method from textual inversion uses a simple prompt that contains only the target object, e.g., ``a photo of $\langle rare \rangle$ dog,'' to generate 64 images. 

Table \ref{tab:result} shows the results. The optimal number of training steps can fluctuate depending on the target objects and their corresponding sample images. As a result, we selected the checkpoints that yielded images most closely matching the training images from among the checkpoints between 200 and 800 steps. For our method, the optimal results emerged from either 200 or 400 steps; for DreamBooth, the best outcomes were between 400 and 800 steps. As shown in Table \ref{tab:result}, our method consistently outperforms DreamBooth across all metrics and testing prompts. Additionally, using simple prompts results in higher scores than using diverse prompts.

\begin{table*}[t]
    \centering
    \label{evaluation}
    \begin{tabular}{ |c|c|c|c|c|  }
    \hline
    \multicolumn{1}{|c|}{} & \multicolumn{2}{|c|}{Diverse Prompt} & \multicolumn{2}{|c|}{Simple Prompt} \\
    \hline
    Method& CLIP score & DINO score & CLIP score & DINO score \\
    \hline
    Ours & 0.800 &0.629 & 0.859 & 0.718\\
    DreamBooth & 0.753   & 0.540  & 0.841 & 0.690\\
    \hline
    \end{tabular}
    \caption{Results of quantitative evaluation.}
    \label{tab:result}
\end{table*}

\subsection{Qualitative Evaluation}

We next show several examples to qualitatively compare the results of the different methods. 

\subsubsection{Quality and Fidelity of Target Object Images}

Figure~\ref{reconstruction} shows a few examples of images generated by the proposed method and DreamBooth for the same set of prompts. The input images for specific target objects are shown on the left (without colored bounding boxes). We input three prompts to our method and DreamBooth, i.e., ``a photo of $\langle rare \rangle$ dog,'' ``a photo of $\langle rare \rangle$ dog on the beach,'' and ``a photo of $\langle rare \rangle$ dog in the jungle.'' The generated images are shown on the right in Figure~\ref{reconstruction}; the results of our method are enclosed by blue boxes and those of DreamBooth are enclosed by red boxes. 

It is first observed that DreamBooth suffers from ``context-appearance entanglement,'' an issue in which the appearance of the target object is influenced by context, as mentioned in \cite{ruiz2023dreambooth}. In DreamBooth's outputs for `backpack,' the color of the backpack changes with the background. This does not occur with our method. This phenomenon may be attributed to the fact that the manually specified identifier inevitably contains some prior knowledge that can affect the generated images in specific contextual settings.
 
The `dog' and `cat' images also show that our method generates images of higher quality and better fidelity.

\begin{figure}
 \centering
 \includegraphics[width=0.9\linewidth]{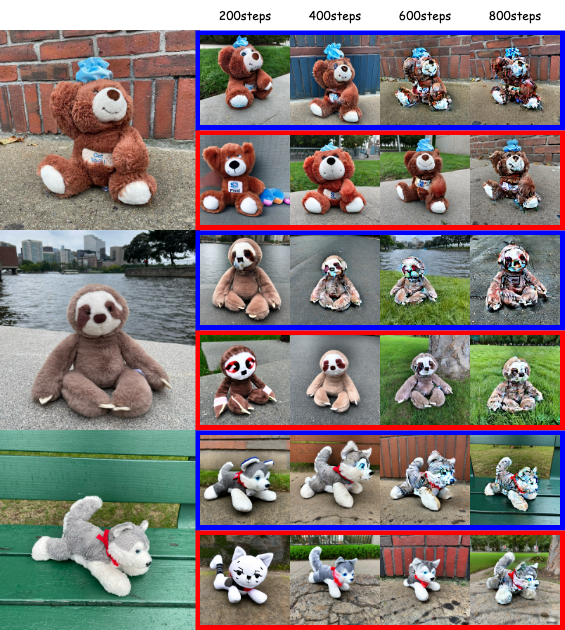}
 \caption{\textbf{Degradation of image quality due to extended training.} Three examples for the object `plushie'
 (stuffed animal). The blue boxes indicate the results of our method using the prompt ``a photo of $\langle rare \rangle$ plushie.'' The red boxes indicate those of DreamBooth using the prompt ``a photo of sks plushie.'' From left to right, the columns correspond to checkpoints at training steps 200, 400, 600, and 800.}
 \label{overtraining}
\end{figure}

\subsubsection{Image Generation for Common Class Objects}

DreamBooth employs the prior preservation loss to mitigate the forgetting of how to generate general object images during training on target object images. While it preserves the diversity of general objects, it may degrade image quality. Since our method requires only a small number of model-parameter updates, the issue of forgetting does not occur even if we do not use this loss function. We demonstrate the impact of the prior preservation loss on general objects by comparing the generated general-object images.

This phenomenon becomes more pronounced as the number of training steps increases. For a fair comparison, we chose checkpoints at the same training step for our method and DreamBooth rather than selecting the checkpoint that performed best on the target object. 
Figure~\ref{priorknowledge} shows a few examples. The images on the left come from two datasets that contain a cat as the target object. The images on the right are generated images guided by the text input ``a photo of cat.'' All the results are from the 200-step checkpoints. It is worth noting that at this step, our method can already reconstruct the target object quite well, while DreamBooth cannot. At the same training step, it is evident that DreamBooth's results (enclosed by red boxes) exhibit a noticeable decrease in the quality of generated general-object images, with a clear difference in realism compared to our approach.

\subsubsection{Longer Training Often Degrades Image Quality}

Training for an extended number of steps frequently results in the generation of distorted or blurry images. This degradation differs from that seen in common-class images, which can suffer severe distortions. In Figure~\ref{overtraining}, the left side displays input sample images, while the right side shows images generated by the models at various training steps. Blue bounding boxes highlight results from our method, while red boxes indicate those from DreamBooth. Notably, not every target object in the dataset exhibits this phenomenon, and the underlying cause remains elusive. We have found that this often happens after the model has robustly learned the target object's features, hinting that overtraining may be responsible. Importantly, our method shows this effect earlier than DreamBooth, suggesting that our method achieves optimal training more efficiently.

\section{Conclusion}

We have proposed a method for personalizing text-to-image diffusion models that enables the model to learn from just four to six sample images of a specific target object. By adding new identifiers, our method can generate images of a specified object in various scenes simply by modifying the input prompt. Our approach produces images that are more similar to the original object while largely preserving the model's inherent capabilities, thereby reducing language drift and image quality degradation. Additionally, our method achieves these enhanced results in significantly less training time.

{\small
\bibliographystyle{ieee_fullname}
\bibliography{personalization}

@misc{ramesh2022dalle2,
      title={Hierarchical Text-Conditional Image Generation with CLIP Latents}, 
      author={Aditya Ramesh and Prafulla Dhariwal and Alex Nichol and Casey Chu and Mark Chen},
      year={2022},
      eprint={2204.06125},
      archivePrefix={arXiv},
      primaryClass={cs.CV}
}

@InProceedings{Rombach_2022_ldm,
    author    = {Rombach, Robin and Blattmann, Andreas and Lorenz, Dominik and Esser, Patrick and Ommer, Bj\"orn},
    title     = {High-Resolution Image Synthesis With Latent Diffusion Models},
    booktitle = {Proceedings of the IEEE/CVF Conference on Computer Vision and Pattern Recognition (CVPR)},
    month     = {June},
    year      = {2022},
    pages     = {10684-10695}
}

@misc{saharia2022imagen,
      title={Photorealistic Text-to-Image Diffusion Models with Deep Language Understanding}, 
      author={Chitwan Saharia and William Chan and Saurabh Saxena and Lala Li and Jay Whang and Emily Denton and Seyed Kamyar Seyed Ghasemipour and Burcu Karagol Ayan and S. Sara Mahdavi and Rapha Gontijo Lopes and Tim Salimans and Jonathan Ho and David J Fleet and Mohammad Norouzi},
      year={2022},
      eprint={2205.11487},
      archivePrefix={arXiv},
      primaryClass={cs.CV}
}

@misc{gal2022textual,
      doi = {10.48550/ARXIV.2208.01618},
      url = {https://arxiv.org/abs/2208.01618},
      author = {Gal, Rinon and Alaluf, Yuval and Atzmon, Yuval and Patashnik, Or and Bermano, Amit H. and Chechik, Gal and Cohen-Or, Daniel},
      title = {An Image is Worth One Word: Personalizing Text-to-Image Generation using Textual Inversion},
      publisher = {arXiv},
      year = {2022},
      primaryClass={cs.CV}
}

@inproceedings{ruiz2023dreambooth,
  title={Dreambooth: Fine tuning text-to-image diffusion models for subject-driven generation},
  author={Ruiz, Nataniel and Li, Yuanzhen and Jampani, Varun and Pritch, Yael and Rubinstein, Michael and Aberman, Kfir},
  booktitle={Proceedings of the IEEE/CVF Conference on Computer Vision and Pattern Recognition},
  year={2023}
}

@misc{ramesh2021dalle,
      title={Zero-Shot Text-to-Image Generation}, 
      author={Aditya Ramesh and Mikhail Pavlov and Gabriel Goh and Scott Gray and Chelsea Voss and Alec Radford and Mark Chen and Ilya Sutskever},
      year={2021},
      eprint={2102.12092},
      archivePrefix={arXiv},
      primaryClass={cs.CV}
}

@misc{yu2022parti,
      title={Scaling Autoregressive Models for Content-Rich Text-to-Image Generation}, 
      author={Jiahui Yu and Yuanzhong Xu and Jing Yu Koh and Thang Luong and Gunjan Baid and Zirui Wang and Vijay Vasudevan and Alexander Ku and Yinfei Yang and Burcu Karagol Ayan and Ben Hutchinson and Wei Han and Zarana Parekh and Xin Li and Han Zhang and Jason Baldridge and Yonghui Wu},
      year={2022},
      eprint={2206.10789},
      archivePrefix={arXiv},
      primaryClass={cs.CV}
}

@article{zhou2021lafite,
  title={LAFITE: Towards Language-Free Training for Text-to-Image Generation},
  author={Zhou, Yufan and Zhang, Ruiyi and Chen, Changyou and Li, Chunyuan and Tensmeyer, Chris and Yu, Tong and Gu, Jiuxiang and Xu, Jinhui and Sun, Tong},
  journal={arXiv preprint arXiv:2111.13792},
  year={2021}
}

@misc{radford2021clip,
      title={Learning Transferable Visual Models From Natural Language Supervision}, 
      author={Alec Radford and Jong Wook Kim and Chris Hallacy and Aditya Ramesh and Gabriel Goh and Sandhini Agarwal and Girish Sastry and Amanda Askell and Pamela Mishkin and Jack Clark and Gretchen Krueger and Ilya Sutskever},
      year={2021},
      eprint={2103.00020},
      archivePrefix={arXiv},
      primaryClass={cs.CV}
}

@article{ho2020ddpm,
  title={Denoising Diffusion Probabilistic Models},
  author={Jonathan Ho and Ajay Jain and Pieter Abbeel},
  year={2020},
  journal={arXiv preprint arxiv:2006.11239}
}

@inproceedings{kumari2022customdiffusion,
  author = {Kumari, Nupur and Zhang, Bingliang and Zhang, Richard and Shechtman, Eli and Zhu, Jun-Yan},
  title = {Multi-Concept Customization of Text-to-Image Diffusion},
  booktitle = {CVPR},
  year = {2023},
}

@misc{huggingface2022dreambooth,
    title = "Training Stable Diffusion with Dreambooth using Diffusers",
    author = "Surak Patil, Pedro Cuenca, Valentine Kozin",
    year = "2022",
    url = "https://huggingface.co/blog/dreambooth"
}

@article{han2023hiper,
      title={Highly Personalized Text Embedding for Image Manipulation by Stable Diffusion},
      author={Han, Inhwa and Yang, Serin and Kwon, Taesung and Ye, Jong Chul},
      journal={arXiv preprint arXiv:2303.08767},
      year={2023}
}

@article{chen2023suti,
        title={Subject-driven Text-to-Image Generation via Apprenticeship Learning}, 
        author={Chen, Wenhu and Hu, Hexiang and Li, Yandong and Ruiz, Nataniel
                and Jia, Xuhui and Chang, Ming-Wei and Cohen, William W},
        journal={arXiv preprint arXiv:2304.00186},
        year={2023},
      }

@article{zhao2023recipe,
  title={A Recipe for Watermarking Diffusion Models},
  author={Zhao, Yunqing and Pang, Tianyu and Du, Chao and Yang, Xiao and Cheung, Ngai-Man and Lin, Min},
  journal={arXiv preprint arXiv:2303.10137},
  year={2023}
}

@inproceedings{kawar2023imagic,
      title={Imagic: Text-Based Real Image Editing with Diffusion Models},
      author={Kawar, Bahjat and Zada, Shiran and Lang, Oran and Tov, Omer and Chang, Huiwen and Dekel, Tali and Mosseri, Inbar and Irani, Michal},
      booktitle={Conference on Computer Vision and Pattern Recognition 2023},
      year={2023}
}

@INPROCEEDINGS{dino,
  author={Caron, Mathilde and Touvron, Hugo and Misra, Ishan and Jegou, Hervé and Mairal, Julien and Bojanowski, Piotr and Joulin, Armand},
  booktitle={2021 IEEE/CVF International Conference on Computer Vision (ICCV)}, 
  title={Emerging Properties in Self-Supervised Vision Transformers}, 
  year={2021},
  volume={},
  number={},
  pages={9630-9640},
  doi={10.1109/ICCV48922.2021.00951}}
}

\end{document}